\documentclass[runningheads]{llncs}
\usepackage[obeyFinal]{easy-todo}
\usepackage{graphicx}
\usepackage{comment}
\usepackage{amsmath,amssymb} 
\usepackage{color}
\usepackage{xspace}
\usepackage{subcaption}
\usepackage{enumitem}
\usepackage{microtype}
\usepackage{wrapfig}
\usepackage{xcolor}
\usepackage{cite}
\usepackage{soul}
\usepackage{booktabs}
\definecolor{citecolor}{HTML}{0071bc}
\usepackage[pagebackref=false,breaklinks=true,letterpaper=true,colorlinks,citecolor=citecolor,bookmarks=false]{hyperref}

\makeatletter
\DeclareRobustCommand\onedot{\futurelet\@let@token\@onedot}
\def\@onedot{\ifx\@let@token.\else.\null\fi\xspace}

\def\eg{\emph{e.g}\onedot} 
\def\ie{\emph{i.e}\onedot}

\def\etal{\emph{et al}\onedot}

\makeatother

\newcommand{\norm}[1]{\left\lVert#1\right\rVert}
\newcommand{\pn}{\textsc{ParSeNet}\xspace}
\newcommand{\sn}{\textsc{SplineNet}\xspace}
\newcommand{\pd}{\textsc{ABCPartsDataset}\xspace}
\newcommand{\sd}{\textsc{SplineDataset}\xspace}
\newcommand{\orcid}[1]{\,\href{https://orcid.org/#1}{\protect\includegraphics[width=8pt]{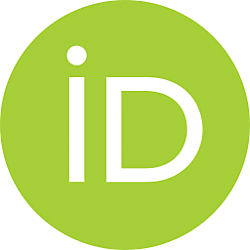}}}
\begin{document}

\pagestyle{headings}
\mainmatter
\title{{\normalfont\textsc{ParSeNet}:} A Parametric Surface Fitting Network for 3D Point Clouds}
\titlerunning{\pn}
\author{Gopal Sharma$^{1}$\orcid{0000-0002-7492-7808}\and
Difan Liu$^{1}$\orcid{0000-0001-5971-2748}\and
Subhransu Maji$^{1}$\orcid{0000-0002-3869-9334}\and \\
Evangelos Kalogerakis$^{1}$\orcid{0000-0002-5867-5735}\and
Siddhartha Chaudhuri$^{2,3}$ \and
Radom\'{\i}r M\v{e}ch$^{2}$
\\\
{\tt\small $^{1}$\{gopalsharma, dliu, smaji, kalo\}@cs.umass.edu, $^{2}$\{sidch,rmech\}@adobe.com}\\
}
\authorrunning{Sharma et al.}
\institute{$^{1}$University of Massachusetts Amherst  $^{2}$Adobe Research  $^{3}$IIT Bombay}
\maketitle

\begin{abstract}
We propose a novel, end-to-end trainable, deep network called \pn that
decomposes a 3D point cloud into parametric surface patches, including
B-spline patches as well as basic geometric primitives. \pn is trained
on a large-scale dataset of man-made 3D shapes and captures high-level
semantic priors for shape decomposition. It handles a much richer
class of primitives than prior work, and allows us to represent
surfaces with higher fidelity. It also produces repeatable and robust
parametrizations of a surface compared to purely geometric
approaches.
We present extensive experiments to validate
our approach against analytical and learning-based alternatives. Our source code is publicly available at: \url{https://hippogriff.github.io/parsenet}. 
\end{abstract}

\newcommand{\ba}{\mathbf{a}}
\newcommand{\bb}{\mathbf{b}}
\newcommand{\bc}{\mathbf{c}}
\newcommand{\bd}{\mathbf{d}}
\newcommand{\be}{\mathbf{e}}
\newcommand{\bff}{\mathbf{f}}
\newcommand{\bg}{\mathbf{g}}
\newcommand{\bh}{\mathbf{h}}
\newcommand{\bi}{\mathbf{i}}
\newcommand{\bj}{\mathbf{j}}
\newcommand{\bk}{\mathbf{k}}
\newcommand{\bl}{\mathbf{l}}
\newcommand{\bm}{\mathbf{m}}
\newcommand{\bn}{\mathbf{n}}
\newcommand{\bo}{\mathbf{o}}
\newcommand{\bp}{\mathbf{p}}
\newcommand{\bq}{\mathbf{q}}
\newcommand{\br}{\mathbf{r}}
\newcommand{\bs}{\mathbf{s}}
\newcommand{\bt}{\mathbf{t}}
\newcommand{\bu}{\mathbf{u}}
\newcommand{\bv}{\mathbf{v}}
\newcommand{\bw}{\mathbf{w}}
\newcommand{\bx}{\mathbf{x}}
\newcommand{\by}{\mathbf{y}}
\newcommand{\bz}{\mathbf{z}}
\newcommand{\bA}{\mathbf{A}}
\newcommand{\bB}{\mathbf{B}}
\newcommand{\bC}{\mathbf{C}}
\newcommand{\bD}{\mathbf{D}}
\newcommand{\bE}{\mathbf{E}}
\newcommand{\bF}{\mathbf{F}}
\newcommand{\bG}{\mathbf{G}}
\newcommand{\bH}{\mathbf{H}}
\newcommand{\bI}{\mathbf{I}}
\newcommand{\bJ}{\mathbf{J}}
\newcommand{\bK}{\mathbf{K}}
\newcommand{\bL}{\mathbf{L}}
\newcommand{\bM}{\mathbf{M}}
\newcommand{\bN}{\mathbf{N}}
\newcommand{\bO}{\mathbf{O}}
\newcommand{\bP}{\mathbf{P}}
\newcommand{\bQ}{\mathbf{Q}}
\newcommand{\bR}{\mathbf{R}}
\newcommand{\bS}{\mathbf{S}}
\newcommand{\bT}{\mathbf{T}}
\newcommand{\bU}{\mathbf{U}}
\newcommand{\bV}{\mathbf{V}}
\newcommand{\bW}{\mathbf{W}}
\newcommand{\bX}{\mathbf{X}}
\newcommand{\bY}{\mathbf{Y}}
\newcommand{\bZ}{\mathbf{Z}}
\newcommand{\balpha}{\mbox{\boldmath$\alpha$}}
\newcommand{\bgamma}{\mbox{\boldmath$\gamma$}}
\newcommand{\bGamma}{\mbox{\boldmath$\Gamma$}}
\newcommand{\bmu}{\mbox{\boldmath$\mu$}}
\newcommand{\bphi}{\mbox{\boldmath$\phi$}}
\newcommand{\bPhi}{\mbox{\boldmath$\Phi$}}
\newcommand{\bSigma}{\mbox{\boldmath$\Sigma$}}
\newcommand{\bsigma}{\mbox{\boldmath$\sigma$}}
\newcommand{\btheta}{\mbox{\boldmath$\theta$}}

\newcommand{\mL}{\mathcal{L}}
\newcommand{\mU}{\mathcal{U}}
\newcommand{\mC}{\mathcal{C}}
\newcommand{\mS}{\mathcal{S}}
\newcommand{\mR}{\mathcal{R}}
\newcommand{\mD}{\mathcal{D}}
\newcommand{\mT}{\mathcal{T}}
\newcommand{\mSl}{\mathcal{S}_l}
\newcommand{\mN}{\mathcal{N}}
\newcommand{\mDll}{\mathcal{D}_{l,l'}}

\newcommand{\ra}{\rightarrow}
\newcommand{\la}{\leftarrow}

\def\A{{\cal A}}
\def\B{{\cal B}}
\def\C{{\cal C}}
\def\D{{\cal D}}
\def\E{{\cal E}}
\def\F{{\cal F}}
\def\G{{\cal G}}
\def\H{{\cal H}}
\def\I{{\cal I}}
\def\J{{\cal J}}
\def\K{{\cal K}}
\def\L{{\cal L}}
\def\M{{\cal M}}
\def\N{{\cal N}}
\def\O{{\cal O}}
\def\P{{\cal P}}
\def\Q{{\cal Q}}
\def\R{{\cal R}}
\def\S{{\cal S}}
\def\T{{\cal T}}
\def\U{{\cal U}}
\def\V{{\cal V}}
\def\W{{\cal W}}
\def\X{{\cal X}}
\def\Y{{\cal Y}}
\def\Z{{\cal Z}}
\def\Re{{\mathbb R}}
\def\Cx{{\mathbb C}}
\def\Ze{{\mathbb Z}}
\def\Na{{\mathbb N}}
\def\ud{\mathrm{d}}
\def\eps{\varepsilon}
\def\dist{\textrm{dist}}

\makeatletter
\renewcommand{\paragraph}{%
  \@startsection{paragraph}{4}%
  {\z@}{1.5ex \@plus 1ex \@minus .2ex}{-0.75em}%
  {\normalfont\normalsize\itshape}%
}
\makeatother

\renewcommand{\labelitemi}{$\bullet$}

\section{Introduction}

\begin{figure}
  \centering
  \includegraphics[width=\linewidth]{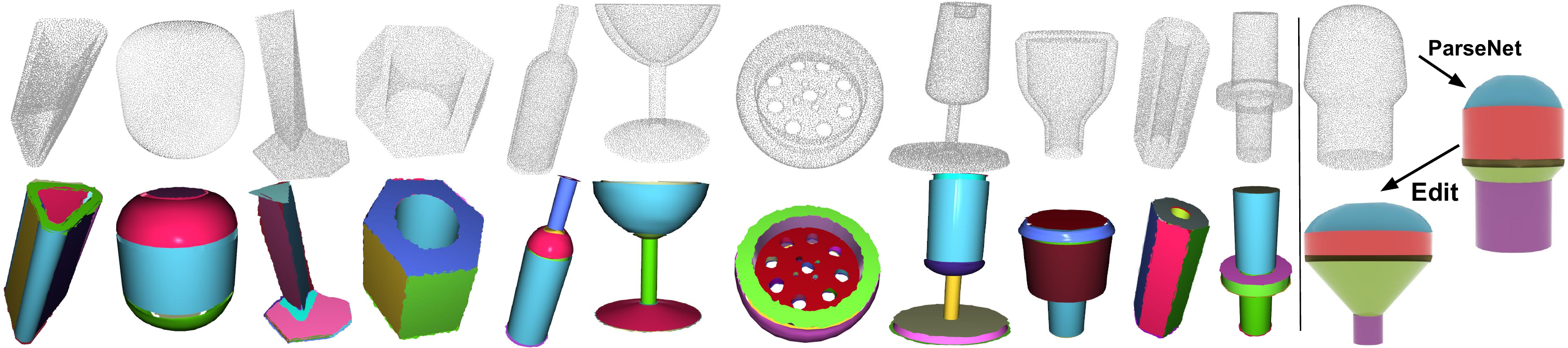}
  \caption{\label{fig:splash} \pn decomposes point clouds (top row) into collections of assembled parametric surface patches including B-spline patches (bottom row). On the right, a shape is edited using the inferred parametrization.}
\end{figure}

3D point clouds can be rapidly acquired using 3D sensors or
photogrammetric techniques. However, they are rarely used in this form
in design and graphics applications. Observations from the
computer-aided design and modeling literature
\cite{Piegl,Schneider,Foley,farin2002curvsurf} suggest that designers
often model shapes by constructing several non-overlapping patches
placed seamlessly. The advantage
of using several patches over a single continuous patch is that a
much more diverse variety of geometric features and
surface topologies can be created. The decomposition also
allows easier interaction and editing.
The goal of this work is to automate the time-consuming process of
converting a 3D point cloud into a piecewise parametric surface
representation as seen in Figure~\ref{fig:splash}.

An important question is how surface patches should be
represented. Patch representations in CAD and graphics are based on
well-accepted geometric properties: (a) \emph{continuity} in their
tangents, normals, and curvature, making patches appear smooth, (b)
\emph{editability}, such that they can easily be modified based on a
few intuitive degrees of freedom (DoFs), \eg, control points or axes,
and (c) \emph{flexibility}, so that a wide variety of surface
geometries can be captured. Towards this goal, we propose \pn, a
\textbf{par}ametric \textbf{s}urfac\textbf{e} fitting \textbf{net}work
architecture which produces a compact, editable representation of a
point cloud as an assembly of geometric primitives, including open or closed B-spline patches.

\begin{figure*}[!t]
  \center{\includegraphics[width=\textwidth]
    {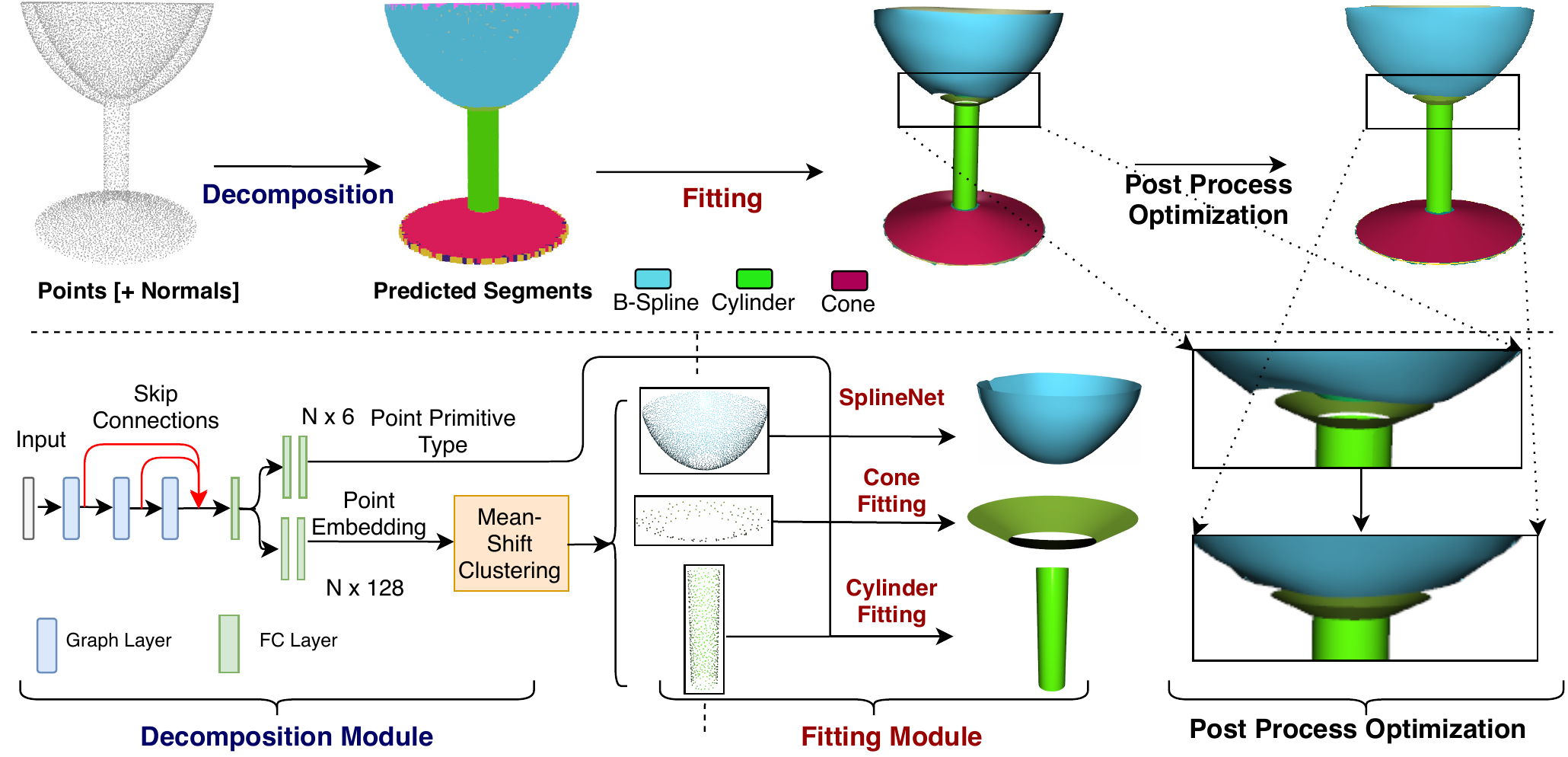}}
\caption{\label{fig:architecture} \textbf{Overview of \textnormal\pn\ pipeline}. (1) The {\em decomposition module} (Section \ref{sec:decomposition-module}) takes a 3D point cloud (with optional normals) and decomposes it into segments labeled by primitive type. (2) The {\em fitting module} (Section~\ref{sec:fitting}) predicts parameters of a primitive that best approximates each segment. It includes a novel \sn to fit B-spline patches. The two modules are jointly trained end-to-end. An optional postprocess module (Section \ref{sec:optimization}) refines the output.}
\end{figure*}

\pn models a richer class of surfaces than prior work which
\emph{only} handles basic geometric primitives such as planes, cuboids
and
cylinders~\cite{tulsiani2017abstraction,PaschalidouSuperquadrics,SharmaCSG,Lingxiao:SPFN}. While
such primitives are continuous and editable representations, they lack
the richness and flexibility of spline patches which are widely used
in shape design. \pn includes a novel neural network (\sn) to estimate
an open or closed B-spline model of a point cloud patch. It is part of
a {\em fitting module} (Section \ref{sec:fitting}) which can also fit
other geometric primitive types. The fitting module receives input from a {\em
decomposition module}, which partitions a point cloud into segments, after which
the fitting module estimates shape parameters of a predicted primitive type for each segment (Section
\ref{sec:decomposition-module}). The entire pipeline, shown in
Figure~\ref{fig:architecture}, is fully differentiable and trained
end-to-end (Section~\ref{sec:training}). An optional geometric
postprocessing step further refines the output.

Compared to purely analytical approaches, \pn produces decompositions
that are more consistent with high-level semantic priors, and are more
robust to point density and noise. To train and test \pn, we leverage
a recent dataset of man-made parts~\cite{Koch2019}. Extensive
evaluations show that \pn outperforms baselines (RANSAC and
SPFN~\cite{Lingxiao:SPFN}) by $14.93\%$ and $13.13\%$ respectively for
segmenting a point cloud into patches, and by $50\%$, and $47.64\%$
relative error respectively for parametrizing each patch for surface reconstruction
(Section~\ref{sec:experiments}).

To summarize, our contributions are:
\begin{itemize}[leftmargin=*]
\item The first proposed end-to-end differentiable approach for representing a raw 3D point cloud as an assembly of parametric primitives {\em including} spline patches.
\item Novel decomposition and primitive fitting modules, including \sn, a fully-differentiable network to fit a cubic B-spline patch to a set of points.
\item Evaluation of our framework vs prior analytical and learning-based methods.
\end{itemize}

\section{Related Work}\label{sec:related-works}
Our work builds upon related research on parametric surface
representations and methods for primitive
fitting. We briefly review relevant work in these areas. Of course, we
also leverage extensive prior work on neural networks for general
shape processing: see recent surveys on the
subject~\cite{ahmed2019dl3d}.

\paragraph{Parametric surfaces.} A parametric surface is a (typically
diffeomorphic) mapping from a (typically compact) subset of $\Re^2$ to
$\Re^3$. While most of the geometric primitives used in computer
graphics (spheres, cuboids, meshes etc) can be represented
parametrically, the term most commonly refers to curved surfaces used
in engineering CAD modelers, represented as spline
patches~\cite{farin2002curvsurf}. There are a variety of formulations
-- \eg B\'ezier patches, B-spline patches, NURBS patches -- with
slightly different characteristics, but they all construct surfaces as
weighted combinations of control parameters, typically the positions
of a sparse grid of points which serve as editing handles.

More specifically, a B-spline patch is a smoothly curved, bounded, parametric surface, whose shape
is defined by a sparse grid of control points $\bC=\{ \bc_{p,q} \}$. The surface
point with parameters \mbox{$(u, v) \in [u_{\min},u_{\max}] \times
  [v_{\min},v_{\max}]$} and {\em basis functions}~\cite{farin2002curvsurf} $b_p(u)$, $b_q(v)$ is given by:
\begin{equation}
\bs(u,v)= \sum\limits_{p=1}^P \sum\limits_{q=1}^Q  b_p(u)b_q(v) \bc_{p,q} \label{eq:b-splines}
\end{equation}
Please refer to supplementary material for more details on B-spline patches.

\paragraph{Fitting geometric primitives.} A variety of analytical (\ie not learning-based) algorithms have been devised to approximate raw 3D data as a collection of geometric primitives: dominant themes include Hough transforms, RANSAC and clustering. The literature is too vast to cover here, we recommend the comprehensive survey of Kaiser \etal\cite{kaiser2019primfit}. In the particular case of NURBS patch fitting, early approaches were based on user interaction or hand-tuned heuristics to extract patches from meshes or point clouds 
\cite{Hoppe94,Eck:1996:ARB,Krishnamurthy:1996:FSS}. In the rest of this section, we briefly review recent methods that {\em learn} how to fit primitives to 3D data.

Several recent
papers~\cite{zou20173d,tulsiani2017abstraction,sun2019abstraction,smirnov2020param}
also try to approximate 3D shapes as unions of cuboids or ellipsoids.
Paschalidou \etal\cite{PaschalidouSuperquadrics,Paschalidou_2020_CVPR} extended
this to superquadrics. Sharma \etal\cite{SharmaCSG,SharmaCSGJournal} developed a
neural parser that represents a test shape as a collection of basic primitives
(spheres, cubes, cylinders) combined with boolean operations. Tian
\etal\cite{tian2018learning} handled more expressive construction rules (\eg.
loops) and a wider set of primitives. Because of the choice of simple
primitives, such models are naturally limited in how well they align to complex
input objects, and offer less flexible and intuitive parametrization for user
edits.

More relevantly to our goal of modeling arbitrary curved surfaces, Gao \etal\cite{Gao:DeepSpline} parametrize 3D point clouds as extrusions or surfaces of revolution, generated by B-spline cross-sections detected by a 2D network. This method requires translational/rotational symmetry, and does not apply to general curved patches. Li \etal\cite{Lingxiao:SPFN} proposed a supervised method to fit primitives to 3D point clouds, first predicting per-point segment labels, primitive types and normals, and then using a differential module to estimate primitive parameters. While we also chain segmentation with fitting in an end-to-end way, we differ from Li \etal in two important ways. First, our differentiable metric-learning segmentation produces improved results (Table \ref{table:segmentation}). Second, a major goal (and technical challenge) for us is to significantly improve expressivity and generality by incorporating B-spline patches: we achieve this with a novel differentiable spline-fitting network.
In a complementary direction, Yumer \etal\cite{yumer2012surface} developed a neural network for fitting a single NURBS patch to an unstructured point cloud. While the goal is similar to our spline-fitting network, it is not combined with a decomposition module that jointly learns how to express a shape with {\em multiple} patches covering different regions. Further, their fitting module has several non-trainable steps which are not obviously differentiable, and hence cannot be used in our pipeline.

\section{Method}\label{sec:method}

The goal of our method is to reconstruct an input point cloud by predicting a set of parametric patches closely approximating its underlying surface.  The first stage of our architecture is a \emph{neural decomposition module} (Fig. \ref{fig:architecture}) whose goal is to segment the input point cloud into regions, each labeled with a parametric patch type. Next, we incorporate a \emph{fitting module} (Fig. \ref{fig:architecture}) that predicts each patch's shape parameters. Finally, an optional post-processing \emph{geometric optimization} step refines the patches to better align their boundaries for a seamless surface.

The input to our pipeline is a set of points $\bP = \{ \bp_i \}_{i=1}^N$,
represented either as 3D\ positions $\bp_i = (x, y, z)$, or as 6D position + normal vectors $\bp_i = (x, y, z, n_x, n_y, n_z)$. The output is a set of surface patches $\{\bs_k\}$, reconstructing the input point cloud. The number of patches is automatically determined. Each patch is labeled with a type $t_k$, one of: sphere, plane, cone, cylinder, open/closed B-spline patch. The architecture also outputs a real-valued vector for each patch defining its geometric parameters, \eg center and radius for spheres, or B-spline control points and knots.

\subsection{Decomposition module}
\label{sec:decomposition-module}
The first module (Fig. \ref{fig:architecture}) decomposes the point cloud $\bP$ into a set of segments such that each segment can be reliably approximated by one of the abovementioned surface patch types. To this end, the module first embeds the input points into a representation space used to reveal such segments. As discussed in Section \ref{sec:training}, the representations are learned using metric learning, such that points belonging to the same patch are embedded close to each other, forming a distinct cluster.

\paragraph{Embedding network.}
To learn these point-wise representations, we incorporate edge convolution layers (EdgeConv) from DGCNN~\cite{Wang:DGCNN}.
Each EdgeConv layer performs a graph convolution to extract a representation of each point with an MLP on the input features of its neighborhood. The neighborhoods are dynamically defined via nearest neighbors in the input feature space.
We stack $3$ EdgeConv layers, each extracting a $256$-D representation per point. A max-pooling layer is also used to extract a global $1024$-D representation for the whole point cloud. The global representation is tiled and concatenated with the  representations from all three EdgeConv layers to form intermediate point-wise $(1024 + 256)$-D representations $\bQ=\{\bq_i\}$ encoding both local and global shape information. We found that a global representation is useful for our task, since it captures the overall geometric shape structure, which is often correlated with the number and type of expected patches. This representation is then transformed through fully connected layers and ReLUs, and finally normalized to unit length to form the point-wise embedding $\bY=\{\by_i\}_{i=1}^N$ ($128$-D) lying on the unit hypersphere.

\paragraph{Clustering.}
A mean-shift clustering procedure is applied on the point-wise embedding to discover segments. The advantage of mean-shift clustering over other alternatives (\eg, k-means or mixture models) is that it does not require the target number of clusters as input. Since different shapes may comprise different numbers of patches, we let mean-shift produce a cluster count tailored for each input. Like the pixel grouping of \cite{kong2018grouppixels}, we implement mean-shift iterations as differentiable recurrent functions, allowing back-propagation. Specifically, we initialize mean-shift by setting all points as seeds $\bz^{(0)}_i=\by_i,  \forall y_i \in R^{128}$. Then, each mean-shift iteration $t$ updates each point's embedding on the unit hypersphere:
\begin{equation}
\bz^{(t+1)}_i = \sum\limits_{j=1}^N \by_j g(\bz_i^{(t)}, \by_j) / (\sum\limits_{j=1}^N g(\bz_i^{(t)}, \by_j))
\label{eq:mean-shift}
\end{equation}
where the pairwise similarities $g(\bz_i^{(t)}, \by_j)$ are based on a von Mises-Fisher kernel with bandwidth $\beta$: $g(\bz_i, \by_j) = \exp(\bz_i^{T} \by_j / \beta^2)$ (iteration index dropped for clarity). The embeddings are normalized to unit vectors after each iteration.
The bandwidth for each input point cloud is set as the average distance of each point to its $150^{th}$ neighboring point in the embedding space~\cite{Silverman86}. The mean-shift iterations are repeated until convergence (this occurs around $50$ iterations in our datasets). We extract the cluster centers using non-maximum suppression: starting with the point with highest density, we remove all points within a distance $\beta$, then repeat. Points are assigned to segments based on their nearest cluster center. The point memberships are stored in a matrix $\bW$, where $\bW[i,k]=1$ means point $i$ belongs to segment $k$, and $0$ means otherwise. The memberships are passed to the fitting module to determine a parametric patch per segment. During training, we use soft memberships for differentiating this step (more details in Section \ref{sec:training_procedure}).

\paragraph{Segment Classification.}
To classify each segment, we pass the per-point representation $\bq_i$, encoding local and global geometry, through fully connected layers and ReLUs, followed by a softmax for a per-point probability $P(t_i = l)$, where $l$ is a patch type (\ie, sphere, plane, cone, cylinder, open/closed B-spline patch). The segment's patch type is determined through majority voting over all its points.

\subsection{Fitting module}\label{sec:fitting}

The second module (Fig. \ref{fig:architecture}) aims to fit a parametric patch to each predicted segment of the point cloud. To this end, depending on the segment type, the module estimates the shape parameters of the surface patch.

\paragraph{Basic primitives.} Following Li \etal\cite{Lingxiao:SPFN}, we estimate the shape of basic primitives with least-squares fitting. This includes center and radius for spheres; normal and offset for planes; center, direction and radius for cylinders; and apex, direction and angle for cones. We also follow their approach to define primitive boundaries.

\paragraph{B-Splines.}

Analytically parametrizing a set of points as a spline patch in the presence of noise, sparsity and non-uniform sampling, can be error-prone. Instead, predicting control points directly with a neural network can provide robust results. We propose a neural network \sn, that inputs points of a segment, and outputs a fixed size control-point grid. A stack of three EdgeConv layers produce point-wise representations concatenated with a global representation extracted from a max-pooling layer (as for decomposition, but weights are not shared). This equips each point $i$ in a segment with a $1024$-D representation $\bphi_i$. A segment's representation is produced by max-pooling over its points, as identified through the membership matrix $\bW$ extracted previously:
\begin{equation}
\bphi_k= \max\limits_{i=1...N} ( \bW[i,k] \cdot \bphi_i).
\label{eq:segment_representation}
\end{equation}
Finally, two fully-connected layers with ReLUs transform $\bphi_k$ to an initial set of $20\times20$ control points $\bC$ unrolled into a 1200-D output vector. For a segment with a small number of points, we upsample the input segment (with nearest neighbor interpolation) to $1600$ points. This significantly improved performance for such segments (Table \ref{table:spline}). For closed B-spline patches, we wrap the first row/column of control points. Note that the network parameters to produce open and closed B-splines are not shared. Fig.~\ref{fig:spline-qual} visualizes some predicted B-spline surfaces.


\subsection{Post-processing module}
\label{sec:optimization}

\sn produces an initial patch surface that approximates the points belonging to a segment. However, patches might not entirely cover the input point cloud, and boundaries between patches are not necessarily well-aligned. Further, the resolution of the initial control point grid ($20 \times 20$) can be further adjusted to match the desired surface resolution. As a post-processing step, we perform an optimization to produce B-spline surfaces that better cover the input point cloud, and refine the control points to achieve a prescribed fitting tolerance.

\paragraph{Optimization.} We first create a grid of $40\times40$ points on the initial B-spline patch by uniformly sampling its UV parameter space. We tessellate them into quads. Then we perform a maximal matching between the quad vertices and the input points of the segment, using the Hungarian algorithm with L2 distance costs. We then perform an as-rigid-as-possible (ARAP) \cite{ARAP_modeling:2007} deformation of the tessellated surface towards the matched input points. ARAP is an iterative, detail-preserving method to deform a mesh so that selected vertices (pivots) achieve targets position, while promoting locally rigid transformations in one-ring neighborhoods (instead of arbitrary ones causing shearing/stretching). We use the boundary vertices of the patch as pivots so that they move close to their matched input points. Thus, we promote coverage of input points by the B-spline patches. After the deformation, the control points are re-estimated with least-squares \cite{Piegl}.

\paragraph{Refinement of B-spline control points.} After the above optimization, we again perform a maximal matching between the quad vertices and the input points of the segment. As a result, the input segment points acquire 2D parameter values in the patch's UV parameter space, which can be used to re-fit any other grid of control points \cite{Piegl}. In our case, we iteratively upsample the control point grid by a factor of $2$ until a fitting tolerance, measured via Chamfer distance, is achieved. If the tolerance is satisfied by the initial control point grid, we can similarly downsample it iteratively. In our experiments, we set the fitting tolerance to $5 \times 10^{-4}$. In Fig. \ref{fig:spline-qual} we show the improvements from the post-processing step.

\section{Training}
\label{sec:training}
To train the neural decomposition and fitting modules of our architecture, we use supervisory signals from a  dataset of
3D shapes modeled through a combination of basic geometric primitives and B-splines.
Below we describe the dataset, then we discuss the loss functions
and  the steps of our training procedure.

\subsection{Dataset}
The ABC\ dataset \cite{Koch2019} provides a large source of 3D CAD models of mechanical objects whose file format stores surface patches and modeling operations that designers used to create them. Since our method is focused on predicting surface patches, and in particular B-spline patches, we selected models from this dataset that contain at least one B-spline surface patch.
As a result, we ended up with a dataset of 32K models (24K, 4K, 4K train, test, validation sets respectively). We call this \pd.
All shapes are centered in the origin and scaled so they lie inside unit cube. To train \sn, we also extract 32K closed and open B-spline surface
patches each from ABC dataset and split them into 24K, 4K, 4K train, test, validation sets
respectively. We call this \sd. We report the average number of different patch types in
supplementary material.

\paragraph{Preprocessing.} Based on the provided metadata in \pd, each shape can be
rendered based on the collection of surface patches and primitives it contains
(Figure \ref{fig:fitting}). Since we assume that the inputs to our architecture
are point clouds, we first sample each shape with 10K points randomly
distributed on the shape surface. We also add noise in a uniform range
$\left[-0.01, 0.01\right]$ along the normal direction. Normals are also
perturbed with random noise in a uniform range of $\left[-3, 3\right]$ degrees
from their original direction.

\begin{figure}[t]
\centering
  \center{\includegraphics[width=0.85\linewidth]
    {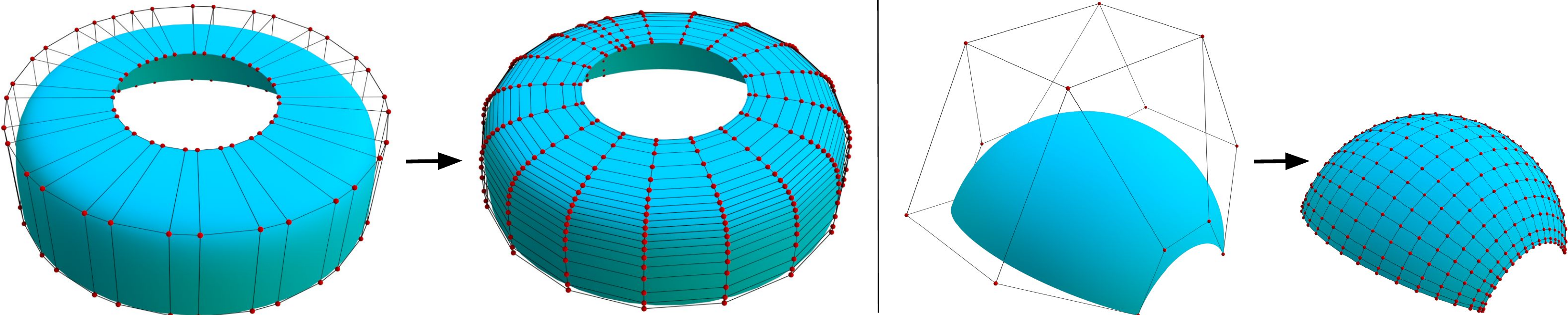}}
  \caption{\label{fig:standard} \textbf{Standardization:} Examples of B-spline patches with a
    variable number of control points (shown in red), each
    standardized with $20\times 20$ control points. Left: closed B-spline and Right: open B-spline. (Please zoom in.)
    }
\end{figure}

\subsection{Loss functions}

\label{sec:losses}
We now describe the different loss functions used to train our neural modules. The training procedure involving their combination is discussed in Section \ref{sec:training_procedure}.

\paragraph{Embedding loss.} To discover clusters of points that correspond well to surface patches, we use a metric learning approach. The point-wise representations $\bZ$ produced  by our decomposition module after mean-shift clustering are learned such that point pairs originating
from the same surface patch \ are embedded close to each other to favor a cluster formation. In contrast, point pairs
originating from different surface patches are pushed away from each other. Given a triplet of points $(a, b, c)$, we use the triplet loss
to learn the embeddings:
c
where $\tau$ the margin is set to $0.9$.
Given a triplet set ${\cal T}_\mS$ sampled from each point set $\mS$ from our dataset $\mD$, the
embedding objective sums the loss over  triplets:
\begin{equation}
L_{emb} = \sum_{\mS \in \mD} \frac{1}{|{\mT}_\mS|} \sum_{(a, b, c) \in {\mT}_\mS} \ell_{emb}(a, b, c).
\label{eq:tripletloss}
\end{equation}
\paragraph{Segment  classification loss.}
To promote correct segment classifications according to our supported types,  we
use the  cross entropy loss:
\mbox{$ L_{class} = -\sum_{i \in \mS} \log(p^{t}_{i})$} where $p^{i}_{t}$ is the probability of the $i^{th}$ point of shape $\mS$ belonging to its ground truth type $t$, computed  from our segment classification network.

\paragraph{Control point regression loss.} This loss function is used to train
\sn. As discussed in Section \ref{sec:fitting}, \sn produces
$20 \times 20$ control points per B-spline patch. We include a
supervisory signal for this control point grid prediction.
One issue is that B-spline patches have a variable number of control
points in our dataset.
Hence we reparametrize each patch by first sampling $M=3600$ points and
estimating a new $20\times 20$ reparametrization using least-squares
fitting~\cite{Piegl,Krishnamurthy:1996:FSS}, as seen in the
Figure~\ref{fig:standard}. In our experiments, we found
that this standardization produces no practical loss in surface reconstructions
in our dataset.
Finally, our reconstruction loss should be invariant to flips or swaps of control points grid in $u$
and $v$ directions.
Hence we define a loss that is invariant to such permutations:
\begin{equation}
  L_{cp} =\sum_{\mS\in \mD}
  \frac{1}{|\mS^{(b)}|}
   \sum_{s_k \in \mS^{(b)}}
  \frac{1}{|\bC_k|}
  \min_{\pi \in \Pi } || \bC_k - \pi( \hat \bC_k ) ||^2
\end{equation}
where $\mS^{(b)}$ is the set of B-spline patches from shape $\mS$, $\bC_k$ is the predicted control point grid for  patch $\bs_k$ ($|\bC_k|=400$ control points),  $\pi( \hat \bC_k )$ is permutations of the
ground-truth control points from the set $\Pi$ of $8$  permutations for open and $160$ permutations for closed B-spline.

\paragraph{Laplacian loss.}  This loss is also specific to
B-Splines using \sn. For each ground-truth B-spline patch, we
uniformly sample ground truth surface, and measure the surface Laplacian
capturing its second-order derivatives. We also uniformly sample
the predicted patches and measure their Laplacians.
We then establish Hungarian matching between sampled points in the ground-truth and predicted
patches, and compare the Laplacians of the ground-truth
points $\hat\br_m$ and corresponding predicted ones
$\br_n$ to improve the agreement between their derivatives as follows:
\begin{equation}
L_{lap} =  \sum_{\mS\in \mD}
  \frac{1}{|\mS^{(b)}| \cdot M}
 \sum_{\bs_k \in \mS^{(b)}} \sum_{\br_n \in s_k }
||\mL(\br_{n}) -  \mL(\hat\br_m)  ||^2
\end{equation}
where $\mL(\cdot)$ is the  Laplace operator on patch points, and $M=1600$ point samples.

\paragraph{Patch distance loss.} This loss is applied to both basic
primitive and B-splines patches.
Inspired by \cite{Lingxiao:SPFN}, the loss measures average distances
between predicted primitive patch $s_k$ and uniformly sampled points
from the ground truth patch as:
\begin{equation}
L_{dist} =  \sum_{\mS\in \mD} \frac{1}{K_{\mS}}\sum_{k=1}^{K_{\mS}} \frac{1}{M_{\hat\bs_k}}\sum_{n \in \hat\bs_k}
D^2( \br_n, \bs_k),
\end{equation}
where $K_\mS$ is the number of predicted patches for shape $\mS$, $M_{\hat\bs_k}$ is number of sampled points $\br_n$ from ground patch $\hat\bs_k$, $D^2( \br_n, \bs_k)$ is the squared distance from $\br_n$ to the predicted primitive patch surface $\bs_k$. These distances can be computed analytically for basic primitives \cite{Lingxiao:SPFN}. For B-splines, we use an approximation based on Chamfer distance between sample points.

\subsection{Training procedure}
\label{sec:training_procedure}
One possibility for training is to start it from scratch using a combination of all losses. Based on our experiments, we found that breaking the training
procedure into the following steps leads to faster convergence  and to
better minima:

\begin{itemize}[leftmargin=*]
\item We  first pre-train the networks of the decomposition module
using \pd with the sum of embedding and classification losses:
$L_{emb}+L_{class}$. Both losses are necessary for point cloud decomposition and classification.

\item We then pre-train the \sn using \sd for control point prediction exclusively on  B-spline patches using $L_{cp}+L_{lap}+L_{dist}$.
We note that we experimented training the B-spline patch prediction
only with the patch distance loss $L_{dist}$ but had  worse
performance. Using both the $L_{cp}$ and $L_{lap}$ loss yielded better
predictions as shown in  Table \ref{table:spline}.

\item We then jointly train the decomposition and fitting module end-to-end
  with all the losses. To allow backpropagation from the
  primitives and B-splines fitting to the embedding network, the mean
  shift clustering is implemented as a recurrent
  module (Equation \ref{eq:mean-shift}). For efficiency, we use $5$
  mean-shift iterations during training.
  It is also important to note that during training,
  Equation \ref{eq:segment_representation} uses \emph{soft}
  point-to-segment memberships, which enables backpropagation from the
  fitting module to the decomposition module and improves reconstructions.
The soft memberships are computed based on the point embeddings
  $\{\bz_i\}$ (after the mean-shift iterations) and  cluster center
  embedding  $\{\bz_k\}$ as follows:
\begin{equation}
  \bW[i,k] = \frac
  {\exp( \bz_k^T \bz_i / \beta^2 ) }
  {\sum_{k'} {\exp( \bz_{k'}^T \bz_i ) / \beta^2 ) } }
\label{eq:membership}
\end{equation}

\end{itemize}

Please see supplementary material for more implementation details.

\section{Experiments}
\label{sec:experiments}

\begin{table*}[t!]
\renewcommand{\arraystretch}{1.1}
  \resizebox{0.9\textwidth}{!}{%
\begin{minipage}{\textwidth}
 \begin{tabular}{l|c|c|c|c|c|c|c}
\toprule
\textbf{Method}               & \textbf{Input} & \textbf{seg iou} & \textbf{label iou} & \textbf{res (all)}         & \textbf{res (geom)}        & \textbf{res (spline)}      & \textbf{P cover}          \\
\hline
NN                   & p     & 54.10   & 61.10     & -                 & -                 & -                 & -                \\
\hline
RANSAC               & p+n   & 67.21   & -         & $0.0220$          & $0.0220$          & -                 & $83.40$          \\
\hline
SPFN                 & p     & 47.38   & 68.92     & $0.0238$          & $0.0270$          & $0.0100$          & $86.66$          \\
SPFN                 & p+n   & 69.01   & 79.94     & $0.0212$          & $0.0240$          & $0.0136$          & $88.40$          \\
\hline
\pn             & p     & 71.32   & 79.61     & $0.0150$          & $0.0160$          & $0.0090$          & $87.00$          \\
\pn             & p+n   & 81.20   & 87.50     & $0.0120$          & $0.0123$          & $0.0077$          & $92.00$          \\
\pn + e2e       & p+n   & 82.14   & 88.60     & $0.0118$          & $0.0120$          & $0.0076$          & $92.30$          \\
\pn + e2e + opt & p+n   & \textbf{82.14}   & \textbf{88.60}     & $\textbf{0.0111}$ & $\textbf{0.0120}$ & $\textbf{0.0068}$ & $\textbf{92.97}$ \\
\bottomrule
 \end{tabular}%
\end{minipage}
  }
  \caption{\label{table:segmentation} \textbf{Primitive fitting on
      \textnormal\pd.}  We compare \pn with nearest neighbor (NN),
    RANSAC~\cite{Schnabel}, and SPFN~\cite{Lingxiao:SPFN}. We show
    results with points (p) and points and normals (p+n) as input. The
    last two rows shows our method with end-to-end training and
    post-process optimization.  We report `seg iou' and `label iou'
    metric for segmentation task. We report the residual error (res)
    on all, geometric and spline primitives, and the coverage
    metric for fitting.}
\end{table*}

Our experiments compare our approach to alternatives in three parts:
(a) evaluation of  the quality of segmentation and segment
classification
(Section \ref{sec:segmentation_evaluation}), (b) evaluation of
 B-spline patch fitting, since it  is a major contribution of our work
 (Section \ref{sec:spline-evaluation}), and (c) evaluation of
 overall reconstruction quality (Section \ref{sec:reconstruction}).
 We
 include evaluation metrics and results for each of the three parts
 next.

\subsection{Segmentation and labeling evaluation}\label{sec:segmentation_evaluation}
\paragraph{Evaluation metrics.} We use the following metrics for evaluating the point cloud segmentation and segment labeling based on the test set of \pd:
\begin{itemize}[leftmargin=*]
\item \textbf{Segmentation mean IOU} (``seg mIOU''): this metric measures the similarity of the predicted segments with ground truth segments. Given the ground-truth point-to-segment memberships $\hat \bW$ for an input point cloud, and the predicted ones $\bW$, we measure:
    $\frac{1}{K}\sum_{k=1}^{K}IOU(\hat\bW[:, k], h(\bW[:, k]))$\\
    where $h$ represents a membership conversion into a one-hot vector, and $K$ is the number of ground-truth segments.

  \item \textbf{Segment labeling IOU} (``label mIOU''): this metric measures the classification accuracy of primitive type prediction averaged over segments:

    $\frac{1}{K} \sum_{k=1}^{K}\mathcal{I}\left[ t_k = \hat{t}_k \right]$
        where $t_k$ and $\hat t_k$ is the predicted and ground truth primitive type respectively for $k^{th}$ segment and $\mathcal{I}$ is an indicator function. 
\end{itemize}
We use Hungarian matching to find correspondences between predicted segments and ground-truth segments.

\begin{figure}[t!]
\begin{centering}
\includegraphics[scale=0.42]{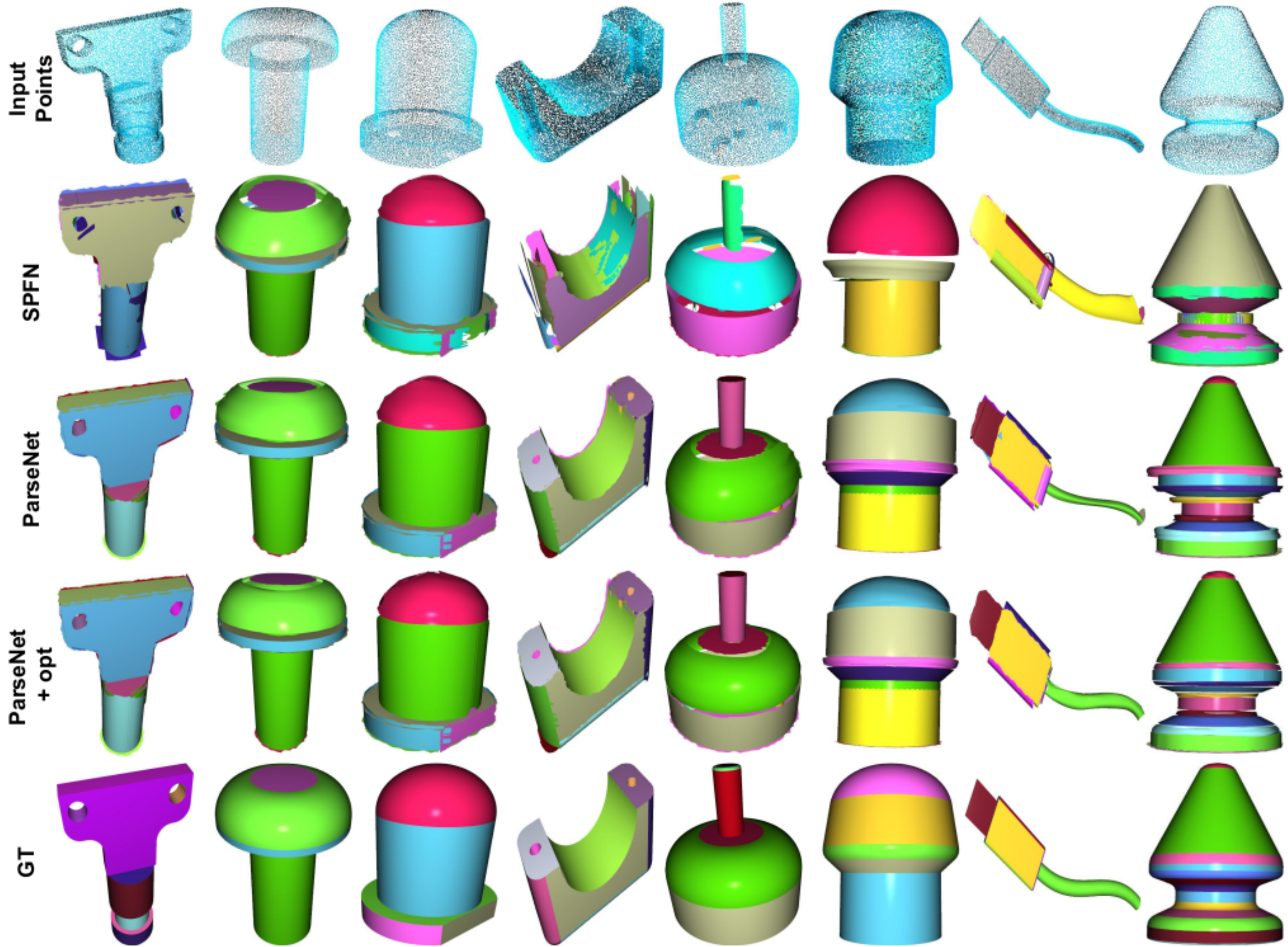}
  \caption{\label{fig:fitting} Given the input point clouds with normals of the first row, we show  surfaces produced by SPFN \cite{Lingxiao:SPFN} (second row), \pn without post-processing optimization (third row), and full \pn including optimization (fourth row). The last row shows the ground-truth surfaces from our \pd.}
\end{centering}
\end{figure}

\paragraph{Comparisons.} We first compare our method with a nearest neighbor (NN) baseline:
for each test shape, we find its most similar shape from the training
set using Chamfer distance. Then for each point on the test shape, we
transfer the labels and primitive type from its closest point in
$\R^3$ on the retrieved shape.

We also compare against efficient RANSAC algorithm
\cite{Schnabel}. The algorithm only handles basic primitives
(cylinder, cone, plane, sphere, and torus), and offers poor
reconstruction of B-splines patches in our
dataset. Efficient RANSAC requires per point normals, which we provide
as the ground-truth normals.
We run RANSAC $3$ times and report the performance with best coverage.

We then compare against the supervised primitive fitting (SPFN)
approach \cite{Lingxiao:SPFN}. Their approach produces per point segment
membership, and their network is trained to maximize relaxed IOU
between predicted membership and ground truth membership, whereas our
approach uses learned point embeddings and clustering with
mean-shift clustering to extract segments.
We train SPFN network using their provided code on our training set
using their proposed losses. We note that we include B-splines patches
in their supported types.
 We train their network in two input settings: (a) the network takes
 only point positions as input, (b) it takes  point and normals as input. We
 train our \pn on our training set in the same two settings using our loss functions.

The performance of the above methods are shown in Table
\ref{table:segmentation}. The lack of B-spline fitting hampers the
performance of RANSAC. The SPFN method with points and normals as
input performs better compared to using only points as input.
Finally, \pn with only points as input performs better than all other
alternatives. We observe further gains when including point normals in
the input. Training \pn end-to-end gives $13.13\%$ and $8.66\%$ improvement in segmentation
mIOU and label mIOU respectively over SPFN with points and normals as input.  The
better performance is also reflected in Figure \ref{fig:fitting},
where our method reconstructs patches that correspond to more
reasonable segmentations compared to other methods.  In the
supplementary material we evaluate methods on
the TraceParts dataset~\cite{Lingxiao:SPFN}, which contains only  basic  primitives (cylinder,  cone,  plane,  sphere, torus). We outperform
prior work also in this dataset.
\begin{table}[t!]
\centering
\renewcommand{\arraystretch}{1.1}
  \setlength{\tabcolsep}{5.5pt}
\resizebox{0.9\textwidth}{!}{%
\begin{minipage}{\textwidth}
\centering
  \begin{tabular}{c|c|c|c|c|c|c|c}
\toprule
\multicolumn{4}{c|}{Loss} & \multicolumn{2}{c|}{Open splines}     & \multicolumn{2}{c}{Closed splines} \\
\hline
cp & dist & lap & opt & \multicolumn{1}{l|}{w/ ups} & w/o ups & w/ ups           & w/o ups          \\ \hline
\checkmark &  &  &                    & $2.04$                      & $2.00$  & $5.04$           & $3.93$           \\
\checkmark
 & \checkmark &  &                & $1.96$                      & $2.00$   & $4.9$            & $3.60$            \\
\checkmark & \checkmark & \checkmark&         & $1.68$                      & $1.59$  & $3.74$           & $3.29$           \\
\checkmark & \checkmark & \checkmark & \checkmark   &  \textbf{0.92}     &  \textbf{0.87}       &    \textbf{0.63} &             \textbf{0.81}
\\
\bottomrule
\end{tabular}
\caption{\label{table:spline}\textbf{Ablation study for B-spline fitting.} The
  error is measured using Chamfer Distance (CD
  is scaled by $100$). The acronyms ``cp'': control-points regression loss, ``dist'' means patch distance loss, and ``lap'' means Laplacian loss. We also include the effect of post-processing optimization ``opt''.  We report performance with and without upsampling (``ups'') for open and closed B-splines.
}
\end{minipage}
}
\end{table}

\begin{figure}[t!]
  \centering
  \includegraphics[scale=0.4]
     {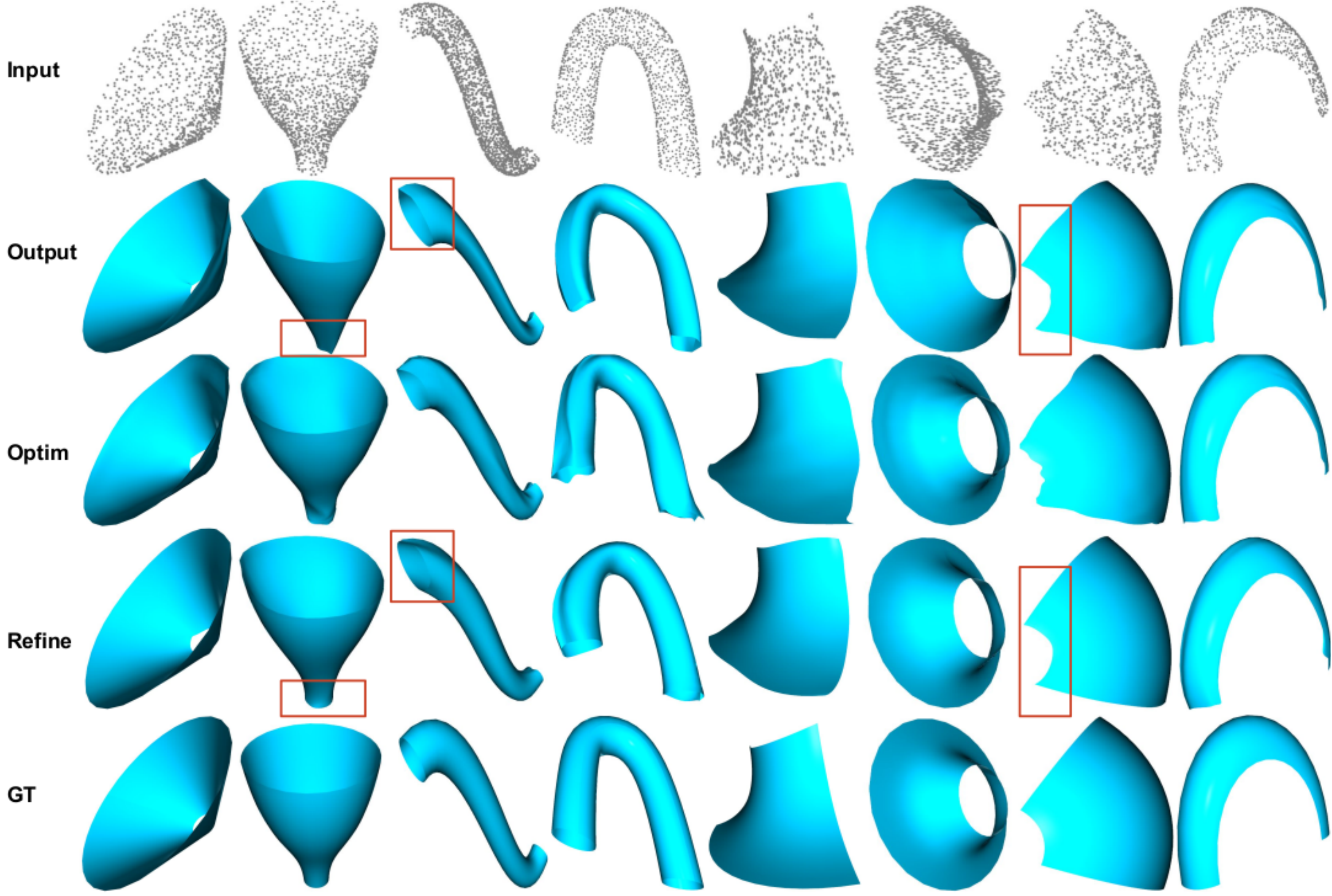}
  \caption{\label{fig:spline-qual}\textbf{Qualitative
      evaluation of B-spline fitting.} From top to bottom: input point cloud, reconstructed surface by \sn,
reconstructed surface by \sn with post-processing optimization, reconstruction
by \sn with control point grid adjustment and finally ground truth surface.
Effect of post process optimization is highlighted in red boxes.}
\end{figure}

\subsection{B-Spline fitting evaluation}\label{sec:spline-evaluation}
\paragraph{Evaluation metrics.} We evaluate the quality of our
predicted B-spline patches by computing the Chamfer distance between
densely sampled points on the ground-truth B-spline patches and
densely sampled points on predicted patches. Points are uniformly
sampled based on the 2D parameter space of the patches. We use 2K
samples. We use the test set of our \sd for evaluation.

\paragraph{Ablation study.}
We evaluate the training of \sn using various loss functions while
giving $700$ points per patch as input, in Table \ref{table:spline}.
All losses contribute to improvements in performance.  Table
\ref{table:spline} shows that upsampling is effective for closed
splines.  Figure \ref{fig:spline-qual} shows the effect of
optimization to improve the alignment of patches and the adjustment
of resolution in the control point grid.
See supplementary material for more experiments on \sn's robustness.

%
\subsection{Reconstruction evaluation}\label{sec:reconstruction}

\paragraph{Evaluation metrics.}
Given a point cloud $\bP = \{ \bp_i\}_{i=1}^N$, ground-truth
patches $\{\cup_{k=1}^{K}\hat \bs_k\}$ and predicted patches
$\{\cup_{k=1}^{K}\bs_k\}$ for a test shape in \pd, we evaluate the
patch-based surface reconstruction using the following:
\begin{itemize}[leftmargin=*]
\item \textbf{Residual error} (``res'') measures the average distance of input
  points from the predicted primitives following \cite{Lingxiao:SPFN}:
  $L_{dist} = \sum_{k=1}^{K} \frac{1}{M_{k}}\sum_{n \in \hat\bs_k } D(\br_n, \bs_k)$
  where $K$ is the number of segments, $M_{k}$ is number of
  sampled points $\br_n$ from ground patch $\hat\bs_k$, $D(\br_n, \bs_k)$ is the distance of $\br_n$ from predicted
  primitive patch $\bs_k$.

\item \textbf{P-coverage} (``P-cover'') measures the coverage of predicted
  surface by the input surface also following \cite{Lingxiao:SPFN}:
  $\frac{1}{N}\sum_{i=1}^{N}\bI\left[\min_{k=1}^{K}D(\bp_i, \bs_k) <
    \epsilon\right]$ ($\epsilon=0.01$).
\end{itemize}

    We note that we use the matched segments after applying
  Hungarian matching algorithm, as in Section \ref{sec:segmentation_evaluation}, to compute these metrics.

\paragraph{Comparisons.} We report the performance of RANSAC for geometric primitive fitting
tasks. Note that RANSAC produces a set of geometric primitives, along
with their primitive type and parameters, which we use to compute the
above metrics. Here we compare with the SPFN network \cite{Lingxiao:SPFN} trained
on our dataset using their proposed loss functions.
We augment their per point primitive type prediction to also include open/closed
B-spline type. Then for classified segments as B-splines, we use our \sn to fit
B-splines. For segments classified as geometric primitives, we use their
geometric primitive fitting algorithm.

\paragraph{Results.} Table \ref{table:segmentation} reports the performance of our method, SPFN and RANSAC. The residual error and
P-coverage follows the trend of segmentation metrics. Interestingly, our method outperforms SPFN even for geometric primitive predictions (even without considering B-splines and our adaptation). Using points and normals, along with joint end-to-end training, and post-processing optimization offers the best performance for our method by giving $47.64\%$ and $50\%$ reduction in relative error in comparison to SPFN and RANSAC respectively.

\section{Conclusion}
We presented a method to reconstruct point clouds by predicting geometric primitives and surface patches common in CAD design.
Our method effectively marries 3D\ deep learning with CAD\ modeling practices.
 Our architecture predictions are editable and interpretable. Modelers can refine our results based on standard CAD modeling operations. 
 In terms of limitations, 
 our method often makes mistakes for small parts,  mainly because clustering merges them with bigger patches.  In high-curvature areas, due to sparse sampling, \pn may produce more segments than ground-truth.
 Producing seamless boundaries is still a challenge due to noise and sparsity in our point sets. Generating training point clouds simulating realistic scan noise is another important future direction.  

\paragraph{Acknowledgements.}
This research is funded in part by NSF (\#1617333, \#1749833) and Adobe.
Our experiments were performed in the UMass GPU cluster funded by the MassTech Collaborative. 
We thank
Matheus Gadelha for helpful discussions.

{\small \bibliographystyle{splncs04}
  \bibliography{egbib} }
\newpage
\def\sm#1{\textcolor{magenta}{{S: }{#1}}}
\section{Supplementary Material}
In our Supplementary Material, we:
\begin{itemize}
  \item provide background on B-spline patches;
  \item provide further details about our dataset,
    architectures and implementation;
  \item evaluate the robustness of \sn as a function of point
    density;
  \item evaluate our approach for reconstruction on the \pd;
  \item show more visualizations of our results; and
  \item evaluate the performance of our approach on the
    TraceParts dataset \cite{Lingxiao:SPFN}.
\end{itemize}

\subsection{Background on B-spline patches.}
A B-spline patch is a smoothly curved, bounded, parametric surface, whose shape is defined by a sparse grid of control points $\bC=\{ \bc_{p,q} \}$. The surface point with parameters \mbox{$(u, v) \in [u_{\min},u_{\max}] \times [v_{\min},v_{\max}]$} is given by:
\begin{equation}
\bs(u,v)= \sum\limits_{p=1}^P \sum\limits_{q=1}^Q  b_p(u)b_q(v) \bc_{p,q} \label{eq:b-splines}
\end{equation}
where $b_p(u)$ and $b_q(v)$ are polynomial B-spline {\em basis
functions}~\cite{farin2002curvsurf}.

To determine how the control points affect the B-spline, a sequence of
parameter values, or {\em knot vector}, is used to divide the range of
each parameter into intervals or {\em knot spans}. Whenever the
parameter value enters a new knot span, a new row (or column) of
control points and associated basis functions become active.
A common knot setting repeats the first and last ones multiple times (specifically
$4$ for cubic B-splines) while keeping the interior knots uniformly
spaced, so that the patch interpolates the corners of the control
point grid. A closed surface is generated by matching the control
points on opposite edges of the grid. There are various
generalizations of B-splines \eg, with rational basis functions or
non-uniform knots. We focus on predicting cubic
B-splines (open or closed) with uniform interior knots, which are
quite common in CAD~\cite{Piegl,Schneider,Foley,farin2002curvsurf}.

\subsection{Dataset}
\begin{figure*}[]
  \begin{subfigure}{0.5\textwidth}
  \label{fig:robust-spline}
  \centering
  \includegraphics[scale=0.35]{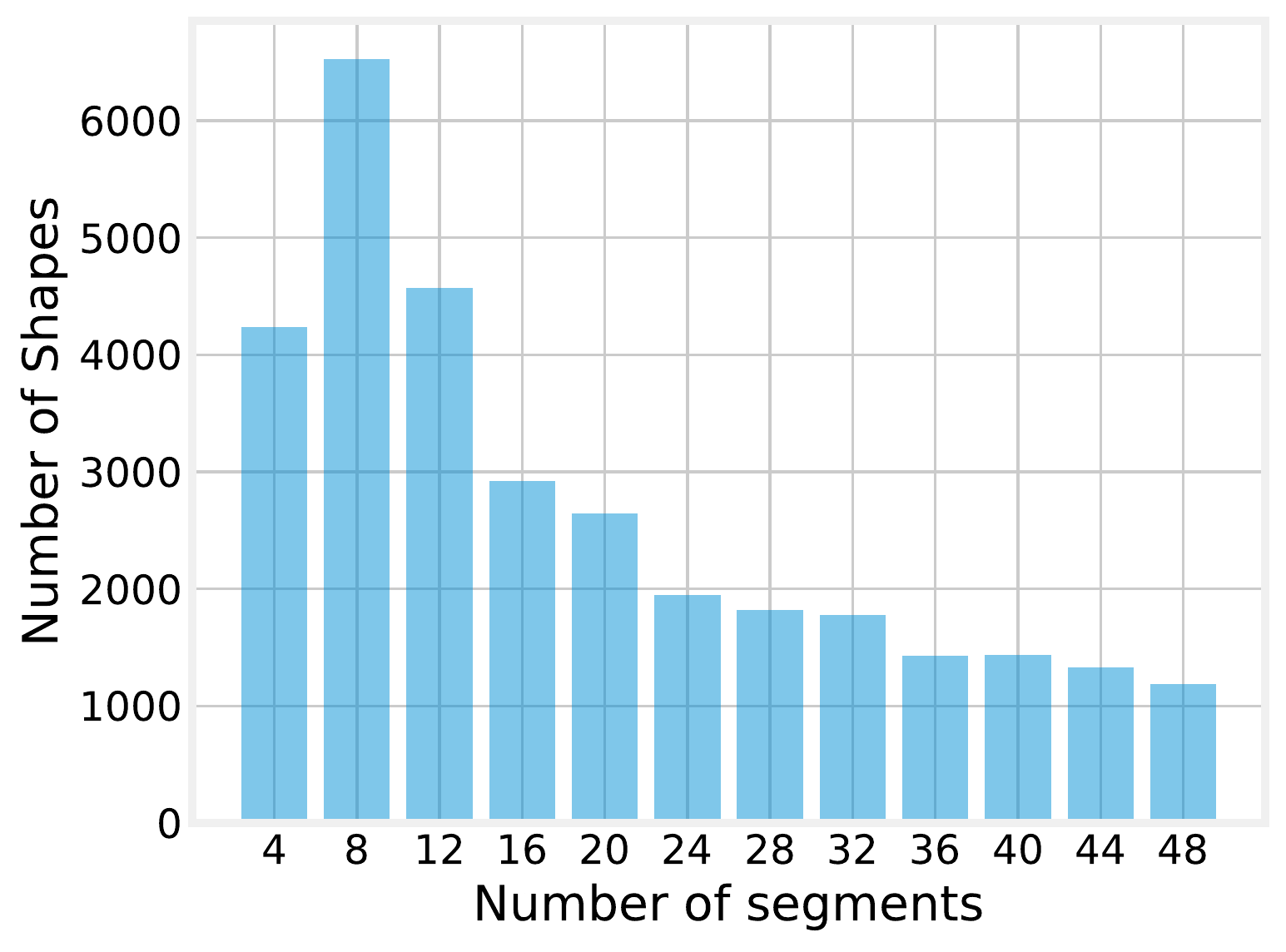}
\end{subfigure}
\begin{subfigure}{0.5\textwidth}
  \label{fig:robust-spline}
  \centering
  \includegraphics[scale=0.35]{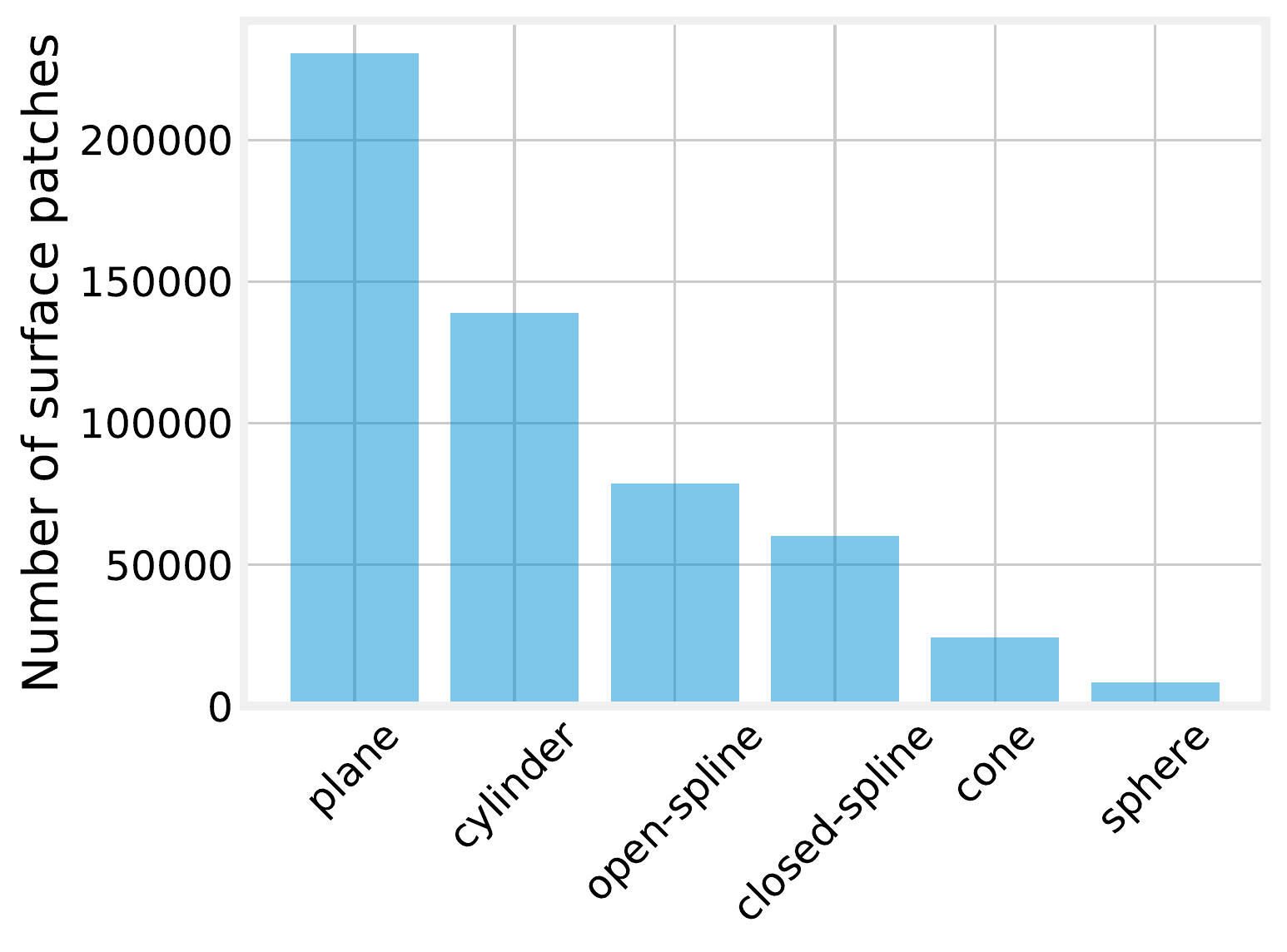}
\end{subfigure}
\caption{\label{fig:stats} \textbf{Histogram of surface patches in
    \textnormal\pd}.
  Left: shows histogram of number of segments and Right:
    shows histogram of primitive types.}
\end{figure*}

The \pd is a subset of the ABC dataset obtained by first selecting
models that contain at least one B-spline surface patch.
To avoid over-segmented shapes, we retain those with up to $50$ surface patches.
This results in a total of $32k$ shapes, which we further split into training
($24k$), validation ($4k$), and test ($4k$) subsets. Figure \ref{fig:stats}
shows the distribution of number and type of surface patches in the dataset.

\subsection{Implementation Details of \textnormal\pn}
\paragraph{Architecture details.} Our decomposition module is based on a dynamic edge convolution network
\cite{Wang:DGCNN}.
The network takes points as input (and optionally normals) and outputs
a per point embedding $Y\in R^{N \times 128}$ and primitive type $T\in
R^{N \times 6}$.  The layers of our network are listed in Table
\ref{table:decomposition-module}.
The edge convolution layer (EdgeConv) takes as input a per-point
feature representation $f \in \mathbb{R}^{N \times D}$,
constructs a kNN graph based on this feature space (we choose $k=80$
neighbors), then forms another feature representation $h \in
\mathbb{R}^{N\times k\times 2D}$, where $h_{i,j} = \left[f_i,
  f_i-f_j\right]$, and $i$,$j$ are neighboring points.
This encodes both unary and pairwise point features, which are further
transformed by a MLP ($D\rightarrow D'$), Group normalization and
LeakyReLU (slope=$0.2$) layers.
This results in a new feature representation: $h' \in
\mathbb{R}^{N\times k \times D'}$.
Features from neighboring points are max-pooled to obtain a per point
feature $f' \in \mathbb{R}^{N\times D'}$.
We express this layer which takes features $f \in \mathbb{R}^{N\times D}$ and
returns features $f' \in \mathbb{R}^{N \times D'}$ as EdgeConv($f$, $D$,
$D'$).
Group normalization in EdgeConv layer allows the use of smaller batch
size during training.
Please refer to \cite{Wang:DGCNN} for more details on edge convolution
network.

\noindent
\sn is also implemented using a dynamic graph CNN.
The network takes points as input and outputs a grid of spline control points
that best approximates the input point cloud.
The architecture of \sn is described in
Table~\ref{table:splinenet-architecture}.
Note that the EdgeConv layer in this network uses batch normalization
instead of group normalization.

\paragraph{Training details.} We use the Adam optimizer for training with learning rate $10^{-2}$ and
reducing it by the factor of two when the validation performance
saturates. For the EdgeConv layers of the decomposition module, we use
$100$ nearest neighbors, and $10$ for the ones in \sn. For
pre-training \sn on \sd, we randomly sample $2k$ points from the
B-spline patches.  Since ABC\ shapes are arbitrarily oriented, we
perform PCA on them and align the direction corresponding to the
smallest eigenvalue to the $+x$ axis. This procedure does not
guarantee alignment, but helps since it reduces the orientation
variability in the dataset. For pre-training the decomposition module
and \sn we augment the training shapes represented as points by using
random jitters, scaling, rotation and point density.

\paragraph{Back propagation through mean-shift clustering.} The $W$
matrix is constructed by first applying non-max suppression (NMS) on
the output of mean shift clustering, which gives us indices of $K$
cluster centers. NMS is done externally \ie outside our
computational graph. We use these indices and Eq. $9$ to compute the $W$
matrix. The derivatives of NMS w.r.t point embeddings are zero or
undefined (i.e. non-differentiable). Thus, we remove NMS from the
computational graph and back-propagate the gradients through a partial
computation graph, which is differentiable. This can be seen as a
straight-through estimator \cite{Schulman2015}. A similar approach is used
in back-propagating gradients through Hungarian Matching in
\cite{Lingxiao:SPFN}. Our experiments in
Table $1$ shows that this approach for end-to-end training is
effective. Constructing a fixed size matrix $W$ will result in
redundant/unused columns because different shapes have different
numbers of clusters. Possible improvements may lie in a continuous
relaxation of clustering similar to differentiable sorting and ranking
\cite{Cuturi2019}, however that is out of scope for our work.

\begin{table}[t]
  \centering \def\arraystretch{1}
\begin{tabular}{c|l|l}
\textbf{Index} & \textbf{Layer}                                            & \textbf{out}       \\
\hline
1 & Input                                                          & N $\times$ 3\\
\hline
2 & EdgeConv(out(1), 3, 64)                                              & N $\times$ 64    \\
\hline
3 & EdgeConv(out(2), 64, 64)                                              & N $\times$ 64    \\
\hline
4 & EdgeConv(out(3), 64, 128)                                             & N $\times$ 128   \\
\hline
5 & CAT(out(2), out(3), out(4))                                   & N $\times$ (256) \\
\hline
6 & RELU(GN(FC(out(5), 1024))) & N $\times$ 1024  \\
\hline
7 & MP(out(6), N, 1)             & 1024      \\
\hline
8 & Repeat(out(7), N)                                                     & N $\times$ 1024  \\
\hline
9 & CAT(out(8), out(5))                                                   & N $\times$ 1280  \\
\hline
10 & RELU(GN(FC(out(9),  512)))                  & N $\times$ 512   \\
\hline
11 & RELU(GN(FC(out(10),  256)))                   & N $\times$ 256   \\
\hline
12 & RELU(GN(FC(out(11),  256)))                   & N $\times$ 256   \\
\hline
13 & \textbf{Embedding}=Norm(FC(out(12),  128))   & N $\times$ 128   \\
\hline
14 & RELU(GN(FC(out(11),  256))                   & N $\times$ 256   \\
\hline
15 & \textbf{Primitive-Type}=Softmax(FC(out(14), 6))      & N $\times$ 6
\end{tabular}
\caption{
  \label{table:decomposition-module}
  \textbf{Architecture of the Decomposition Module.}
  EdgeConv: edge convolution, GN: group
  normalization, RELU: rectified linear unit, FC: fully connected
  layer, CAT: concatenate tensors along the second dimension, MP:
  max-pooling along the first dimension, Norm: normalizing the
  tensor to unit Euclidean length across the second dimension.}

\end{table}

\begin{table}[]
\centering
\begin{tabular}{l|l|l}
\textbf{Index} & \textbf{Layer}                        & \textbf{Output}     \\
\hline
1 & Input                                      & N $\times$ 3 \\
\hline
2 & EdgeConv(out(1),3, 128)                              & N $\times$ 128    \\
\hline
3 & EdgeConv(out(2),128, 128)                             & N $\times$ 128    \\
\hline
4 & EdgeConv(out(3),128, 256)                             & N $\times$ 256    \\
\hline
5 & EdgeConv(out(4),256, 512)                             & N $\times$ 512    \\
\hline
6 & CAT(out(2), out(3), out(4), out(5)))                                   & N $\times$ (1152) \\
\hline
7 & RELU(BN(FC(out(6), 1024)) & N $\times$ 1024   \\
\hline
8 & MP(out(7), N, 1)                & 1024       \\
\hline
9 & RELU(BN(FC(out(8), 1024))                            & 1024       \\
\hline
10 & RELU(BN(FC(out(9), 1024))                            & 1024       \\
\hline
11 & Tanh(FC(out(10), 1200))                               & 1200       \\
\hline
12 & \textbf{Control Points} = Reshape(out(11), (20, 20, 3))                                           & 20 $\times$ 20 $\times$ 3
\end{tabular}
\caption{\label{table:splinenet-architecture}\textbf{Architecture of
    \textnormal\sn.} EdgeConv: edge convolution layer, BN: batch noramlization,
  RELU: rectified linear unit, FC: fully connected layer, CAT:
  concatenate tensors along second dimension, and MP: max-pooling
  across first dimension}
\end{table}

\subsection{Robustness analysis of \sn}
\begin{figure*}[t!]
  \begin{subfigure}{0.5\textwidth}
  \label{fig:robust-spline}
  \centering
  \includegraphics[scale=0.35]{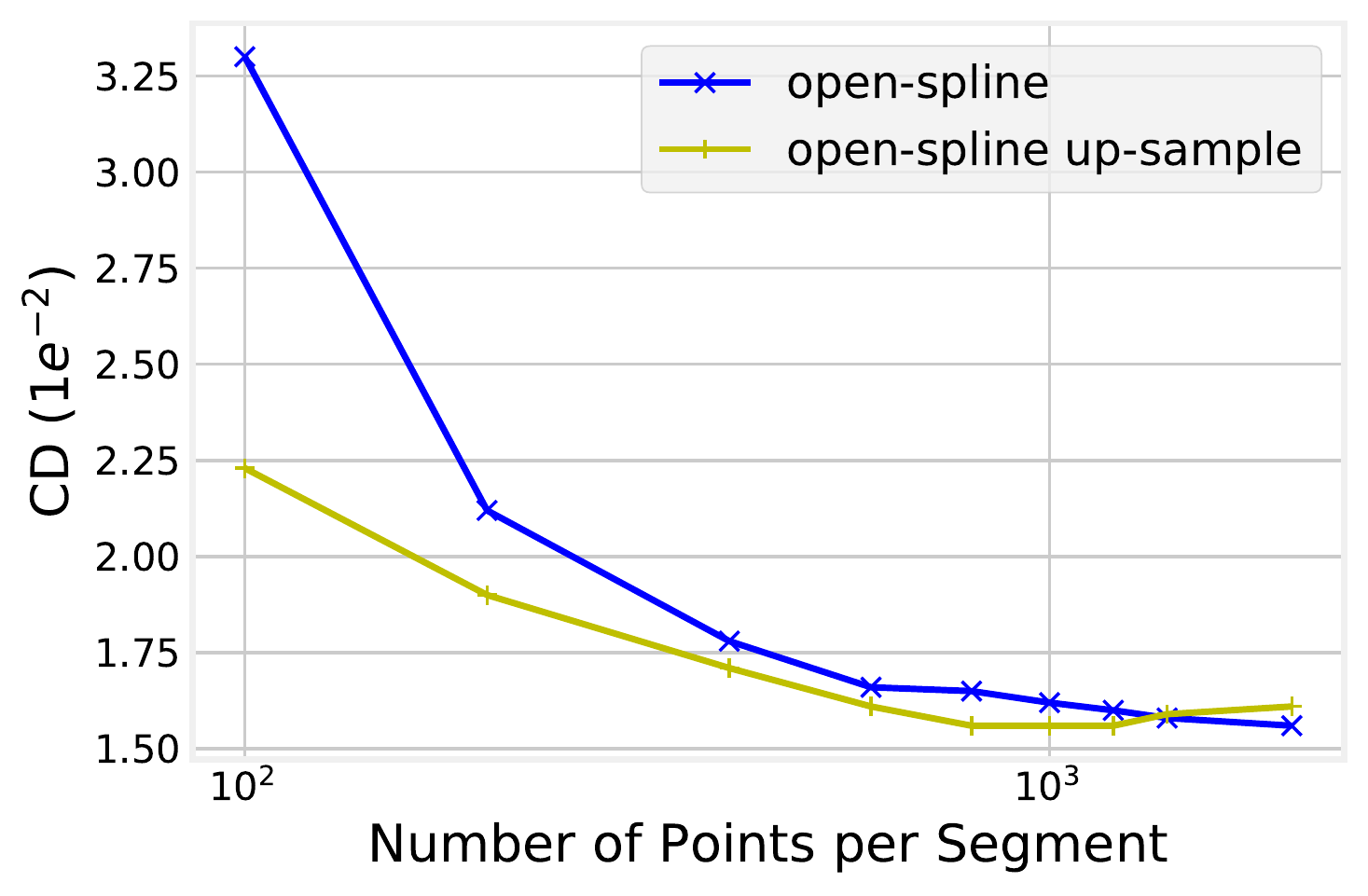}
\end{subfigure}
\begin{subfigure}{0.5\textwidth}
  \label{fig:robust-spline}
  \centering
  \includegraphics[scale=0.35]{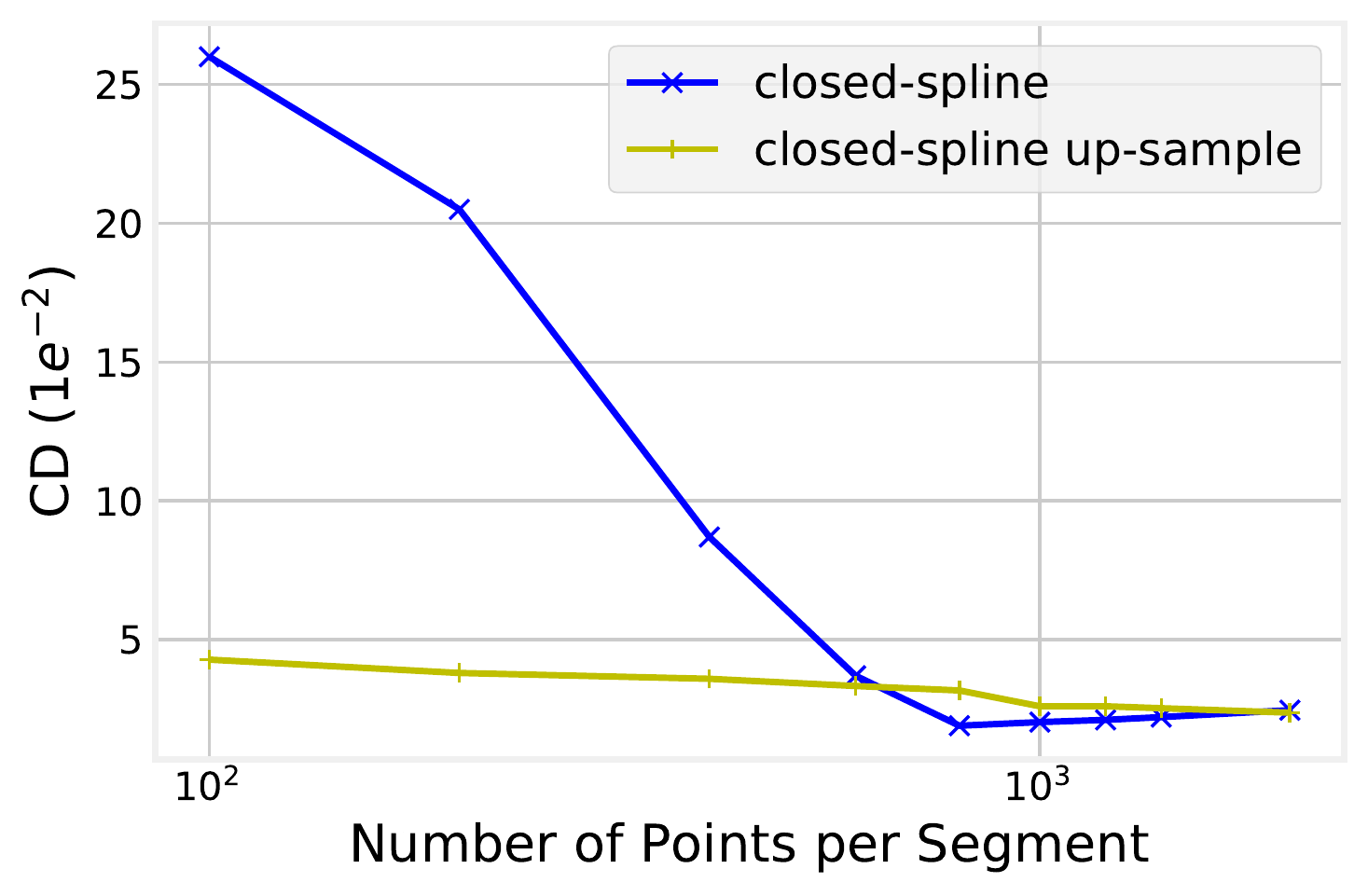}
\end{subfigure}
\caption{\label{fig:robust-spline} \textbf{Robustness analysis
  of \textnormal\sn.} Left: open B-spline and Right: closed B-spline. Performance
  degrades for sparse inputs (blue curve).  Nearest neighbor
  up-sampling of the input point cloud to $1.6K$ points reduces error for
  sparser inputs (yellow curve). The horizontal axis is in log scale.
  The error is measured using Chamfer distance (CD).}
\label{fig:robust-spline}
\end{figure*}

Here we evaluate the performance of \sn as a function of the point
sampling density. As seen in Figure~\ref{fig:robust-spline}, the
performance of \sn is low when the point density is small ($100$
points per surface patch).
\sn is based on graph edge convolutions \cite{Wang:DGCNN}, which
are affected by the underlying sampling density of the network.
However, upsampling points using a nearest neighbor
interpolation leads to a significantly better performance.

\subsection{Evaluation of Reconstruction using Chamfer Distance}
Here we evaluate the performance of \pn and other
baselines for the task of reconstruction using Chamfer distance on
\pd.
Chamfer distance between
reconstructed points $P$ and input points $\hat{P}$ is defined as:
\begin{equation*}\label{eq:s-cover}
  p_{cover} = \frac{1}{|P|} \sum_{i\in P}\min_{j \in \hat{P}}\norm{i-j}^2,
\end{equation*}

\begin{equation*}\label{eq:p-cover}
  s_{cover} = \frac{1}{|\hat{P}|} \sum_{i\in \hat{P}}\min_{j \in P}\norm{i-j}^2,
\end{equation*}
\begin{equation*}\label{eq:cd-eval}
  CD = \frac{1}{2}(p_{cover} + s_{cover}).
\end{equation*}
Here $|P|$ and $|\hat{P}|$ denote the cardinality of $P$ and $\hat{P}$
respectively. We randomly sample $10k$ points each on the predicted and ground
truth surface for the evaluation of all methods. Each predicted surface patch is
also trimmed to define its boundary using bit-mapping with epsilon $0.1$
\cite{Schnabel}. To evaluate this metric, we use all predicted
surface patches instead of the \emph{matched} surface patches that is used in
Section $5.3$.

\begin{figure}
  \centering
  \includegraphics[width=\linewidth]{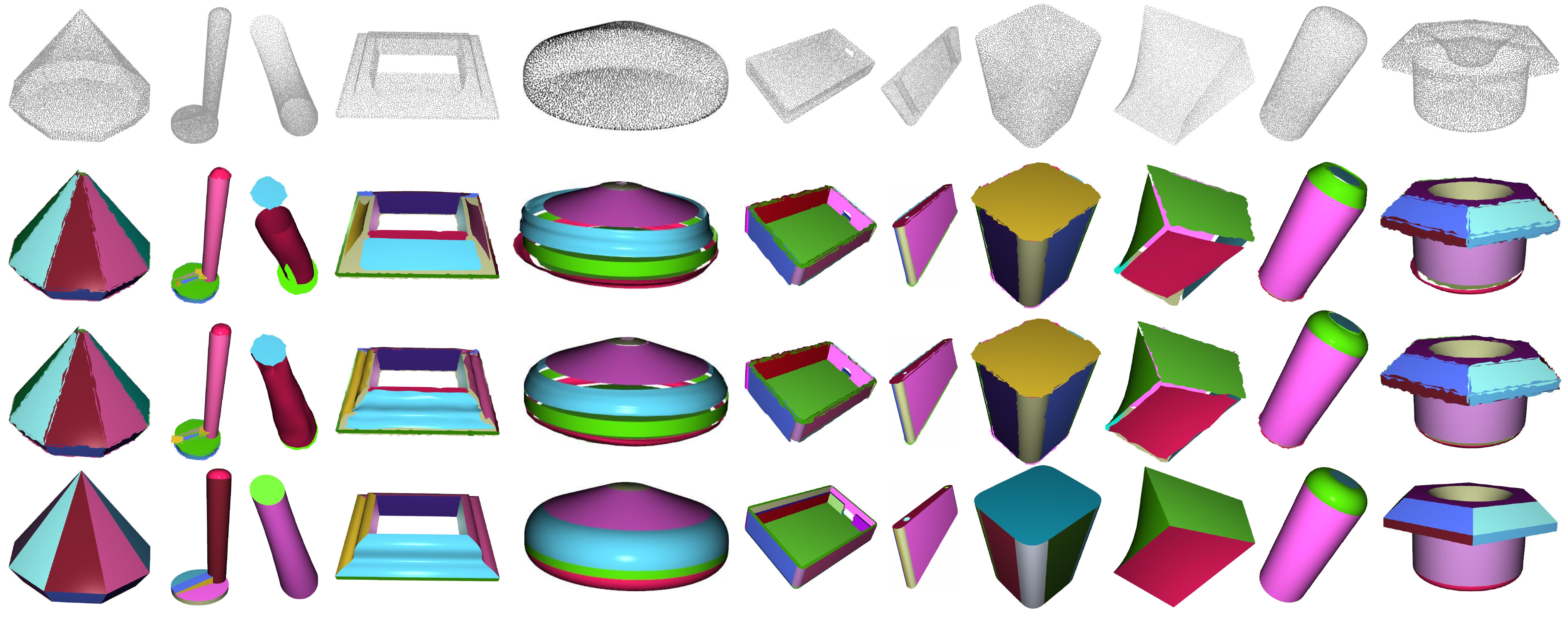}
  \caption{\label{fig:supplementary-fitting} Given the input point clouds with normals in the first row, we show  surfaces produced by \pn without post-processing optimization (second row), and full \pn including optimization (third row). The last row shows the ground-truth surfaces from our \pd.}
\end{figure}

Results are shown in Table \ref{table:fitting-chamfer-distance}. Evaluation using Chamfer
distance follows the same trend of residual error shown in Table $1$. \pn and SPFN with points as input performs better than NN
and RANSAC. \pn and SPFN with points along with normals as input performs better
than with just points as input. By training \pn end-to-end and also using post-process optimization results in the best performance. Our full \pn gives
$35.67\%$ and $49.53\%$ reduction in relative error in comparison to SPFN and
RANSAC respectively. We show more visualizations of surfaces reconstructed by \pn in Figure
\ref{fig:supplementary-fitting}.

\begin{table}[t]
  \centering
  \begin{tabular}{l|c|c|c|c}
    \toprule
  \textbf{Method}                 & \textbf{Input} & \textbf{p cover ($1\times 10^{-4}$)} & \textbf{s cover ($1\times 10^{-4}$)} & \textbf{CD ($1\times 10^{-4}$)}\\
  \hline
  NN                     & p   & $10.10$   & $12.30$        &  $11.20$                    \\
  \hline
  RANSAC                 & p+n & $7.87$   &  $17.90$       &  $12.90$                    \\
  \hline
  SPFN                   & p   & $7.17$   &  $13.40$       &  $10.30$                    \\
  SPFN                   & p+n & $6.98$   &  $13.30$       &  $10.12$                    \\
  \hline
  \pn                    & p   & $6.07$   &  $12.40$       &  $9.26$                     \\
  \pn                    & p+n & $4.77$   &  $11.60$       &  $8.20$                     \\
  \pn + e2e + opt        & p+n & $\textbf{2.45}$      &  $\textbf{10.60}$       &   $\textbf{6.51}$                        \\
  \bottomrule
\end{tabular}
  \caption{\label{table:fitting-chamfer-distance} \textbf{Reconstruction error measured using Chamfer distance on \textnormal\pd.} `e2e': end-to-end training of \pn and `opt': post-process optimization applied to B-spline surface patches.}
\end{table}

\subsection{Evaluation on TraceParts Dataset}
Here we evaluate the performance of \pn on the TraceParts dataset, and compare
it with SPFN. Note that the input points are normalized to
lie inside a unit cube.
Points sampled from the shapes in TraceParts
\cite{Lingxiao:SPFN} have a fraction of points not assigned to any
cluster.
To make this dataset compatible with our evaluation approach, each
unassigned point is merged to its closest cluster. This results in evaluation score to differ from the reported score in their paper \cite{Lingxiao:SPFN}.

First we create a nearest neighbor (NN) baseline as shown in the Section
$5.3$. In this, we first scale both training and testing shape an-isotropically such that each dimension has unit length. Then for each test shape, we find its most similar shape from the training
set using Chamfer Distance. Then for each point on the test shape, we
transfer the labels and primitive type from its closest point in
$\R^3$ on the retrieved shape.  We train \pn on the training set of
TraceParts using the losses proposed in the Section $4.2$
and we also train SPFN using their proposed losses.
All results are reported on the test set of TraceParts.

\begin{figure}
  \centering \includegraphics[width=\linewidth]{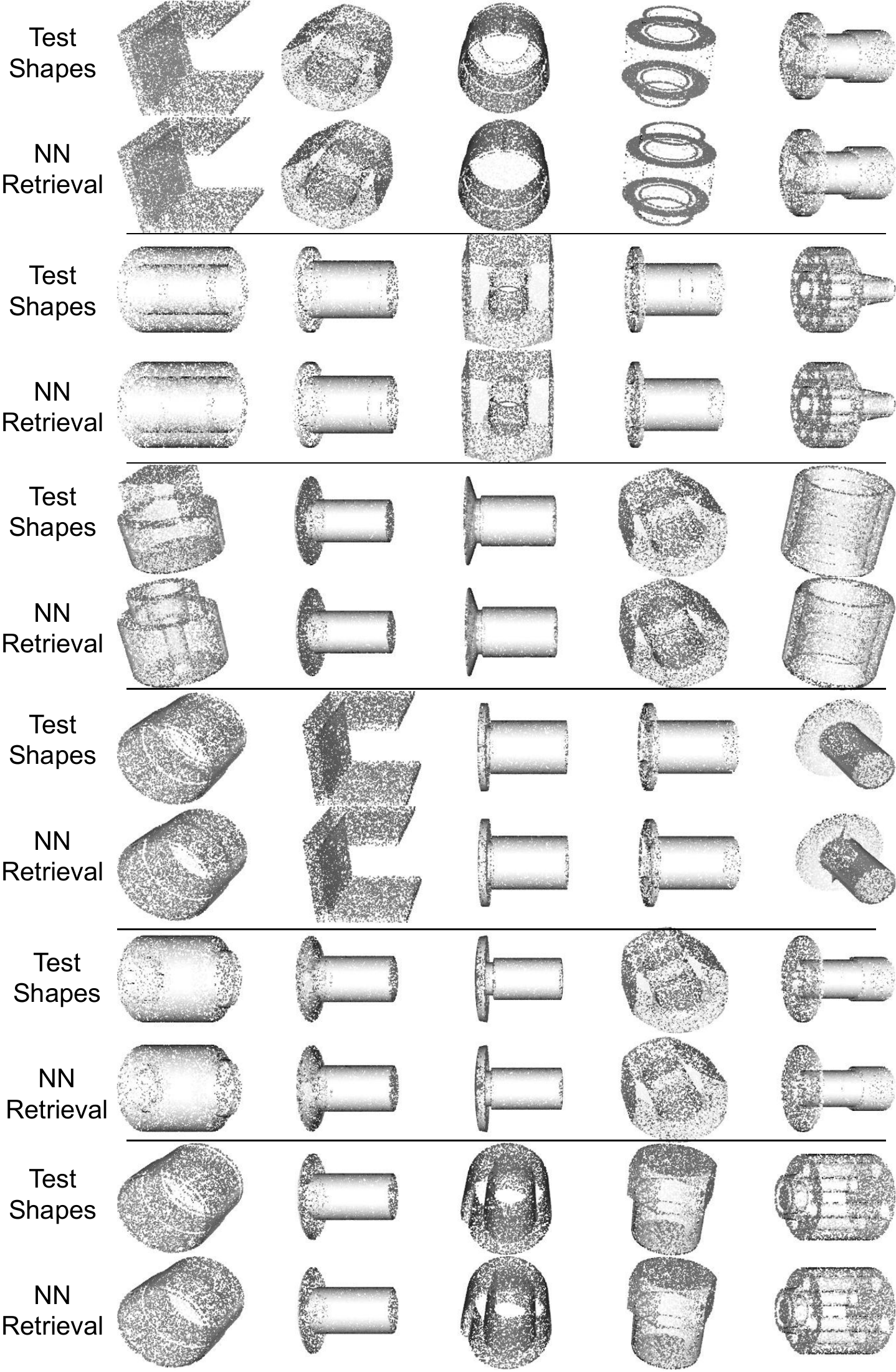} \caption{\label{fig:traceparts-nn} \textbf{Nearest
  neighbor retrieval on the TracePart dataset} We randomly select 30
  shapes from the test set of TraceParts dataset and show the NN
  retrieval, which reveals high training and testing set overlap. Shapes are
  an-isotropically scaled to unit length in each dimension. This is further validated quantitatively in
  Table \ref{table:TraceParts-seg}.}
\end{figure}

Results are shown in the Table \ref{table:TraceParts-seg}. The NN approach achieves
a high segmentation mIOU of $81.92\%$ and primitive type mIOU of $95\%$.
Figure~\ref{fig:traceparts-nn} shows the NN results for a random set of shapes
in the test set. It seems that the test and training sets often contain
duplicate or near-duplicate shapes in the TraceParts dataset. Thus the
performance of the NN can be attributed to the lack of shape diversity in this
dataset. In comparison, our dataset is diverse, both in terms of shape variety
and primitive types, and the NN baseline achieve much lower performance with
segmentation mIOU of $54.10\%$ and primitive type mIOU of $61.10\%$.

We further compare our \pn with SPFN with just points as input.
\pn achieves $79.91\%$ seg mIOU compared to $76.4\%$ in SPFN.
\pn achieves $97.39\%$ label mIOU compared to $95.18\%$ in SPFN.
We also perform better when both points and normals are used as input to \pn and
SPFN.

Finally, we compare reconstruction performance in the Table
\ref{table:TraceParts-recon}. With just points as input to the
network, \pn reduces the relative residual error by $9.35\%$ with
respect to SPFN. With both points and normals as input \pn reduces
relative residual error by $15.17\%$ with respect to SPFN.

\begin{table}[]
  \centering
  \begin{tabular}{c|c|c|c}
      \toprule
    \textbf{Method}   & \textbf{Input} & \textbf{seg mIOU} & \textbf{label mIOU} \\
    \hline
    NN       & p     & 81.92   & 95.00 \\
    \hline
    SPFN     & p     & 76.4    & 95.18 \\
    SPFN     & p + n & 88.05   & 98.10 \\
    \hline
    ParseNet & p     & 79.91   & 97.39 \\
    ParseNet & p + n & \textbf{88.57}   & \textbf{98.26} \\
      \bottomrule
  \end{tabular}
  \caption{\label{table:TraceParts-seg} \textbf{Segmentation results on the TraceParts dataset.} We report segmentation and primitive type prediction performance of various methods.}
\end{table}

\begin{table}[]
  \centering
  \begin{tabular}{c|c|c|c}
      \toprule
    \textbf{Method}   & \textbf{Input} & \textbf{res} & \textbf{P cover} \\
    \hline
    NN       & p     & 0.0138  & 91.90 \\
    \hline
    SPFN     & p     & 0.0139    & 91.70   \\
    SPFN     & p + n     & 0.0112    & \textbf{92.94}   \\
    \hline
    ParseNet & p     & 0.0126   & 90.90    \\
    ParseNet & p + n & 0\textbf{.0095}   & 92.72    \\
    \bottomrule
  \end{tabular}
  \caption{
    \label{table:TraceParts-recon}
    \textbf{Reconstruction results on the TraceParts dataset.} We report residual loss and P cover metrics for various methods.
  }
\end{table}

\end{document}